# Urdu Morphology, Orthography and Lexicon Extraction

**Muhammad Humayoun**
Department of Mathematics
University of Savoie
France
mhuma@univ-savoie.fr

**Harald Hammarström, Aarne Ranta**
Department of Computer Science
Chalmers University of Technology
Sweden
{harald2,aarne}@cs.chalmers.se

## Abstract

Urdu is a challenging language because of, first, its Perso-Arabic script and second, its morphological system having inherent grammatical forms and vocabulary of Arabic, Persian and the native languages of South Asia. This paper describes an implementation of the Urdu language as a software API, and we deal with orthography, morphology and the extraction of the lexicon. The morphology is implemented in a toolkit called Functional Morphology (Forsberg & Ranta, 2004), which is based on the idea of dealing grammars as software libraries. Therefore this implementation could be reused in applications such as intelligent search of keywords, language training and infrastructure for syntax. We also present an implementation of a small part of Urdu syntax to demonstrate this reusability.

## 1 Introduction

Urdu is an Indo-Aryan language widely spoken in Pakistan, India, Jammu & Kashmir and Bangladesh. It is also spoken in all over the world due to the big South Asian Diaspora. Urdu is closely related to Hindi and it shares morphology, syntax and almost all phonology.

However the two languages differ in their script, some of the phonology and vocabulary. Urdu has a strong Perso-Arabic influence in its vocabulary and is written in a cursive, context-sensitive Perso-Arabic script from right to left; whereas Hindi has a strong influence of Sanskrit and is written in the Devanagari script from left to right.

Despite the differences, both languages share grammar (morphology + syntax) and a huge amount of vocabulary. Urdu-Hindi together is the second most widely spoken language in the world with 1,017,290,000 speakers (Native + second language) after Chinese (Rahman, 2004).

Today the state of the art technology to write morphologies is to use special-purpose languages based on finite-state technology. The most well-known is XFST (Xerox Finite State Tool) which is based on regular expressions. In our opinion, these languages are still close to the machine code. Therefore, we emphasis on using a higher level language to capture the linguistic abstraction. Then, that higher level code should be translated into finite state code by some tool if required.

## 2 Goals and Contributions

Implementing grammar for Urdu also requires dealing with orthography. In this paper we present the process of creating the following resources:

1) The morphology component as an open source software API having:
   i) A type system that covers the language abstraction of Urdu completely.
   ii) An inflection engine that covers word-and-paradigm morphological rules for Urdu for every word class.
   iii) Rules for automatic lexicon extraction using the *extract* tool (Forsberg, Hammarström & Ranta, 2006)
   iv) A lexicon of 4,816 words.
2) An orthography component containing Unicode Infrastructure for the Urdu morphology API to accommodate Perso-Arabic script of Urdu, including a GUI application and useful tools.
3) A syntax component containing the implementation of a small part of Urdu syntax in Grammatical Framework (Ranta, 2004) by reusing the above mentioned components.

The overall picture of the Urdu morphology is shown in the following diagram:

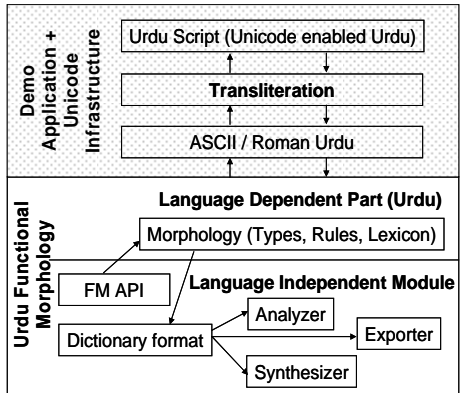

**Figure 1**: Urdu Functional Morphology

## 3 Urdu Orthography and Unicode Infrastructure

Urdu has an alphabet of 57 letters (Afzal & Hussain, 2001, p 2) and 15 diacritic marks (Aərɑb/Hərkɑt̪). Urdu orthography inherits some characteristics from Arabic such as the optional use of diacritic marks. Similarly the short vowels are not considered as letters of their own but applied above or below a consonant by using appropriate diacritics.

In this work, it is decided to store all the work related to Urdu morphology in ASCII characters so that:

- It could be viewed and manipulated easily on different platforms.
- The same inflection engine could be reused for Hindi morphology in future, just by adding a lexicon and the transliteration scheme for Hindi.

Since we want the end product to support Unicode Character set, therefore a clear, strict and reversible transliteration scheme is defined both for the letters and the diacritic marks of Urdu script. This transliteration-function is one-to-one i.e. one string in Urdu gives exactly one roman string and vice versa (but one Urdu character need not correspond to exactly one roman character). It takes Urdu script (represented by Unicode Character codes) or Roman transliteration script and transliterates it in the opposite system of writing (Urdu ↔ Roman). In a similar way the corresponding phonetic string can also be retrieved from the Roman or Urdu transliteration unambiguously. It is implemented in Java by using a Transliterator class of ICU4J (ICU4J 3.4, 2006) which is an open source API for software internationalization.

The roman correspondences to the Urdu letters were mostly chosen to be reminiscent of prototypical pronunciation values of the ASCII characters. However, precedence is given to the widely used keys in Urdu keyboard layouts and to the characters used in roman Urdu on Internet. This makes the transliteration easy-to-adapt for Urdu users.

Following are some example words converted from Urdu script to equivalent Roman transliteration:

|   | Urdu | Meaning | Roman |
|---|------|---------|-------|
| 1 | کوشش (koʃɪʃ) | Struggle | k(a)wX(i)X |
| 2 | بھاگ (bʰɑg) | Run | b|hag |

**Table 1**: Transliteration examples

Other than the Urdu Transliteration component, the following four utilities are developed as a part of the Unicode infrastructure:

- The Urdu Unicode Extractor
- The Urdu Keyboard Input Method
- The Main GUI application

The *Urdu Extractor* is developed to extract the Urdu Unicode text from web pages, and has further been used for the collection of corpus (section 6).

An onscreen *Urdu keyboard* is developed so that a user can type Urdu without installing a specific Urdu keyboard. To render and display Urdu correctly, an Urdu Font has been embedded inside this tool.

The above mentioned tools are then combined and are further used to build a *main GUI application* that interface the Functional Morphology (henceforth FM) runtime system into Java to provide morphological analysis, both in Urdu and Roman. We interface four kinds of analyses which are provided by FM as a part of its runtime system.

## 4 Functional Morphology Toolkit

FM is a toolkit for morphology development in Haskell (Forsberg & Ranta, 2004). It is based on an idea of using the high expressiveness provided by functional languages to define morphology. The use of Haskell gives access to powerful programming constructs and high level of abstraction, which is very useful to capture generalizations about a natural language.

FM toolkit is a successful experiment of how the morphology can be implemented in Haskell by using it as a host language while FM acts as a domain specific embedded language. The productivity and reliability of FM has already been proved by successful implementations of the morphologies for Swedish, Italian, Russian, Spanish and Latin (FM, 2004).

### 4.1 Functional Morphology overview

FM consists of two parts, first, the language independent part and, second, the language dependent part as shown in Figure 1.

A morphology implementer takes the language independent part as granted. It consists of, first, an infrastructure for dictionary compilation, second, the runtime applications (Analyzer, synthesizer) and third, the data export utility (Translator). The Translator can export morphology in many formats such as XFST & LEXC, SQL database and Grammatical Framework (henceforth GF) grammar source code etc, providing wider usability aspects. FM consists of three main type classes: *Param, Dict* and *Language*. These type classes enable code reuse and provide generic algorithms for analysis, synthesis and code generation.

The language dependent part of the system is the one that a morphology implementer has to provide for a certain language which is Urdu in our case. The implementation then appears as a new library on top of the language independent part of FM. The language dependent part consists, first a type system, second an inflection engine and third a lexicon.

## 5 Urdu Morphology

For morphology, Urdu is quite similar to other Indo-European languages, e.g. having concatenative inflective morphological system. However some differences can be found in case of causative verbs that also exhibit stem-internal changes in some cases. In this section we discuss the Urdu morphology and then present our solution that explains them in FM. In morphology, we do not deal with the words that would require determining across multi-token units (some loan derivational affixes), hence leaving them to be dealt at the syntax level.

### 5.1 Nouns

In Urdu, nouns are inflected for number and case, and have an inherent gender that could be either masculine or feminine. Urdu has three cases for nouns (Schmidt, 1999, p 7); the nominative, oblique and vocative. Following are the types defined for case and number:

data Case = Nom | Obl | Voc
data Number = Sg | Pl

The inflectional types, *Case* and *Number* are then combined as a type named *NounForm*, which is further used to define a type *Noun.*

data NounForm = NF Number Case
type Noun = NounForm → Str

All types are language dependent and to be able to use the common API functions, they should be the valid instances of *Param* class. We do it in a following way:

instance Param Case where values       = enum
instance Param Number where values     = enum
instance Param NounForm where
        values = [NF n c| n←values, c←values]

Nouns can be divided into different groups based on their inflection. We divide nouns on the basis of the ending letters in their singular forms. We started our work by making suitable divisions of nouns into groups which are mentioned as groups by (Siddiqi, p 287, 289, 302-304) and (Schmidt, 1999, p 4). However, changes have been done to group the words with respect to pure morphological perspective. This resulted into fifteen groups including three groups for loan Arabic nouns and two for loan Persian nouns. We show one group below as a running example:

- Singular masculine nouns ending with (ا, a), (ه, h) and (ع, e):

This group also includes the Arabic loan nouns ending with (ه, h). According to this rule, first, if a word ends with letter (ا, a) or (ه, h) then:

- To make plural nominative and singular oblique, the last letter is replaced by letter (ے, E)
- To make plural oblique, the last letter is replaced by string (وں, wN) and

- To make plural vocative, the last letter is replaced by letter ( و, w)

Second, if a word ends with ( ع, e) then the rules will remain same as above except that the above mentioned letters will be added at the end of words without replacing any existing letter.

Following is a table displaying forms of this group e.g. (ləɽka, لَڑکا, boy).

|  | Nominative | Oblique | Vocative |
|---|---|---|---|
| Singular | ləɽka لَڑکا | ləɽke لَڑکے | ləɽke لَڑکے |
| Plural | ləɽke لَڑکے | ləɽkoɳ لَڑکوں | ləɽko لَڑکو |

**Table 2**: An example noun group

This group is defined in inflection engine in the following way:

noun_lRka :: DictForm → Noun
noun_lRka lRka nf =
  mkNoun sg pl pl pl_obl sg_obl pl_voc nf
    where
        sg       = lRka
        pl       = lRk ++ "E"
        pl_obl   = lRk ++ "wN"
        pl_voc   = lRk ++ "w"
        lRk      = if (end =="e") then lRka
                    else (tk 1 lRka)
        end      = dp 1 lRka

This function generates the appropriate forms for different cases and then passes them to a more generic function (*mkNoun*) as parameters. It is defined in the following way:

mkNoun:: String → String → String → String →
  String → String → Number → Case → String
mkNoun sg sg_Obl pl pl_Obl sg_Voc pl_Voc n c =
  case (n,c) of
    (Sg, Nom)    → sg
    (Sg, Obl)    → sg_Obl
    (Sg, Voc)    → sg_Voc
    (Pl, Nom)    → pl
    (Pl, Obl)    → pl_Obl
    (Pl, Voc)    → pl_Voc

Then an interface function for this group is defined in *BuildUrdu.hs*, which is a coordinator between type system, Inflection engine and the lexicon, in the flowing way:

n1 :: DictForm → Entry
n1 df = masculine (noun_lRka df)

*DictForm* is string type, and *masculine* is a function applied on such functions that are written for the inflection of masculine words. Then this interface function *n1* is added in *CommandsUrdu.hs* to let it behave like a command for the lexicon and words are added in the lexicon in the following way:

n1 l(a)R'ka   (transliteration of ləɽka, لَڑکا )

## 5.2 Verbs

The Urdu verbs are complex as compared to the other word classes. Urdu verb inflects for tense, mood, aspect, gender and number. Many verb auxiliaries are used to represent the tense, mood and aspect of a verb. Furthermore these auxiliaries also inflect as normal verbs.

Urdu verb shows direct and indirect causative behavior. In general, for each verb, there exists at least one stem form that could be Intransitive, transitive etc. This basic stem form then normally makes two other forms (direct & indirect causatives) of that verb. These three forms are actually regular verbs. They make conjugation independently and can have similar or different meanings from each other. For example, consider a verb (bən, بن, be made):

|  | Root | Infinitive | Oblique |
|---|---|---|---|
| Intransitive / (di) Transitive | bən بن | bənna بننا | bənne بننے |
| Direct Causative | bəna بنا | bənana بنانا | bənane بنانے |
| Indirect Causative | bənwa بنوا | bənwana بنوانا | bənwane بنوانے |

**Table 3:** An Example verb

(bənna, بننا, be made), (bənana, بنانا, to make/cause to make) and (bənwana, بنوانا, cause to be made) in Table 3 are masculine infinitive forms of three regular verbs.

### 5.2.1 Verb categories

In the perspective of morphology, we divide verbs in the following categories:

1) Verbs only having basic stem form, while direct & indirect causatives do not exist
2) Verbs having basic stem form as well as direct & indirect causatives. The direct and indirect causatives are made by:
   i) Rules
   ii) Irregulars

3) Verbs only having basic and direct causative forms, while indirect causative does not exist.
4) Verbs only having basic and indirect causative forms, while direct causative does not exist

Morphologically, in Urdu, a verb inflects in:

- Gender (Masculine, Feminine)
- Number (Singular, Plural)
- Person (First, Second {casual, familiar, respectful}, Third)
- Mood & Tense (Subjunctive, Perfective, Imperfective)

We show the implementation of the second category as running example. We define it in the type system in the following way:

```
type Verb = VerbForm → Str
data VerbForm =
   VF Tense Person Number Gender          |
   Caus1 Tense Person Number Gender       |
   Caus2 Tense Person Number Gender       |
   Inf          | Caus1_Inf     | Caus2_Inf       |
   Inf_Fem      | Caus1_Inf_Fem | Caus2_Inf_Fem   |
   Inf_Obl      | Caus1_Inf_Obl | Caus2_Inf_Obl   |
   Root         | Caus1_Root    | Caus2_Root

data Tense = Subj | Perf | Imperf
```

The first three constructors in the definition of *VerbForm* provide verb analysis for the basic verb form, direct and indirect causative forms respectively. Similarly *Inf, Caus1_Inf* and *Caus2_Inf* are infinitive masculine forms; *Inf_Fem, Caus1_Inf_Fem, Caus2_Inf_Fem* are the feminine infinitive forms; *Inf_Obl, Caus1_Inf_Obl* and *Caus2_Inf_Obl* are infinitive oblique forms; and *Root, Caus1_Root, Caus2_Root* are root forms.

As described above, the verbs belonging to this category could be formed by rules; however there exist a big number of irregular verbs as well. We show the worst-case function written for irregular verbs:

```
mkVerbCaus12 :: String -> String -> String -> Verb
mkVerbCaus12 vInf caus1_inf caus2_inf =
  mkGenVerb root r1 r2 vInf caus1_inf caus2_inf
       where
          root   = (tk 2 vInf)
          r1     = (tk 2 caus1_inf)
          r2     = (tk 2 caus2_inf)
```

In this function we provide the basic, direct and indirect causative forms as arguments. As in Urdu, the conjugation of verbs is very regular; a complete inflection can be built from these three forms. An interface function v4 is defined for this group and they are added in lexicon as follows:

v4 m(i)l'na m(i)lana m(i)l'wana (مِلنا ، مِلانا ، مِلوانا)

A general function *mkGenVerb* is used to produce the complete verb conjugation with morphological point of view.

## 5.3 Adjectives, Adverbs and closed classes

In a similar way the Adjectives, Adverbs, Pronouns, Postpositions, Particles and Numerals have been implemented with similar level of detail.

## 6 The Lexicon

A wide-coverage lexicon is a key part of any morphological implementation. Today, most of the lexicons are built manually, which is a very time consuming task. We aim to build a lexicon automatically with minimal human effort. We use a tool named extract which is primarily designed for the morphologies developed in FM.

A morphology implementer provides a *paradigm* file and a corpus to the tool. The tool reads the rules for all paradigms, searches the corpus for those words that fulfill the definition of paradigms and extracts them along with the name of the fulfilled paradigm.

For the extraction of Urdu lexicon, the first step was to collect a reasonable amount of Urdu Unicode text to make a corpus. We developed an Urdu corpus of 1.5 million words from news and literature domain (book banks and news on Internet). It was tokenized on space and punctuation marks, keeping the diacritics. It returned 63,700 unique words.

It is interesting to note that the unique words are considerably less (23.86 times less) then the total words in the corpus. This conforms well to our intuition that high frequent items, such as postpositions, auxiliaries, particles and pronouns, account for most tokens in Urdu text.

We devised 26 rules (6 for verbs, 19 for nouns, 1 for adjectives) to write a paradigm file for Urdu. Let's look at the rule defined for irregular verbs that has basic, direct & indirect causative forms:

paradigm v4 = x +"na" x+"ana" x+"wana"
    { x+"na" & (x+"ana" | x+"wana") };

It results the output in a following format that is saved directly in the lexicon:

v4 dyk|hna d(i)k|hana d(i)k|hwana (دیکھنا، دِکھانا، دِکھوانا)

Then the tool is applied on the corpus along with the paradigm file, resulting in an Urdu lexicon of 9,126 words. This result could vary with respect to, first, the occurrence of misspellings, foreign words, numeric expressions, pronouns etc in the corpus; second, the knowledge of the lexical distribution of the language; and third, the level of strictness in the paradigm rules. Here strictness means a tighter definition of a paradigm rule by requiring more word forms as a condition.

Like most of the other Arabic script-based languages, Urdu is commonly written without or with a variant number of diacritic marks in electronic and print media. It specifically appears as a fundamental limitation to get a fully vocalized corpus to build a lexicon. This situation may also lead to a problem of having more versions per word with different diacritic information in automatic extraction of the lexicon. For example, for a word kɪt̪ɑb, we may get (کِتاب, kɪt̪ɑb) and (کتاب, kt̪ɑb) in the lexicon which are two orthographically different words representing the same word. Therefore in such cases, it is most desirable to save only one version of such words in the lexicon with full diacritics.

However, the words with different diacritic information are not always the same words. They may be different in their meanings; e.g. (تَیر, t̪ær, to swim) and (تِیر, t̪ɪr, arrow). In such cases, it is important to save all such words in the lexicon with full diacritics.

Further, since the use of Urdu on Internet is relatively new, we were also expecting a relatively high number of spelling mistakes in the extracted corpus. Therefore, to be sure about the correctness of the lexicon with respect to the points raised above, we manually re-checked the lexicon from word to word; and all incorrect entries have been thrown away resulting in a lexicon of 4,816 words, generating 137,182 word forms. However, we did not apply the missing diacritics on partly vocalized words which could be seen as a fundamental limitation of our lexicon.

The manually checked lexicon (4,816 words) is approximately half (52.8%) of the extracted lexicon (9,126 words). We found that the incorrect entries are mostly due to, first, the spelling mistakes; second, the lack of spaces between words or extra spaces inside words; and third, the use of foreign words; e.g. the use of Arabic and Persian text in Urdu, mostly in the religious, as well as in the slightly old literary text; where text is normally aided by the Quranic verses and the Persian poetry. Similarly, the text from news domain shows a big number of proper nouns and foreign words taken from English.

Following are the results altogether:

|  | Words | Diacritic words |
|---|---|---|
| Corpus | 1,520,000 | 23,696 |
| Unique | 63,700 | 6,633 |
| Extracted lexicon | 9,126 | 632 |
| Clean lexicon | 4,816 | 415 |

**Table 3: Results**

## 7 Urdu Syntax

Despite the fact that Urdu is an Indo-European language, its syntax shows many differences from the other Indo-European languages due to the inherent features of Arabic, Persian and the native languages of the Indo-Pak subcontinent. The pragmatically neutral constituent order in Urdu is SOV (Subject Object Verb).

To show the usability and effectiveness of our approach, we provide an implementation of a small part of Urdu syntax in GF which is an open source special-purpose programming language for defining grammars.

We port the Urdu morphology API from FM to GF by using data export utility of FM. Later we apply some preprocessing and save the lexicon directly in Unicode Urdu for GF and then we build the syntax as a separate part of the system on top of the morphology.

In GF, a grammar is a combination of two parts: The Abstract syntax and the Concrete syntax. Below we show some functions of the Abstract syntax from our implementation:

fun UsePastS: NP → VP → S;
fun UsePresS: NP → VP → S;

In our implementation, a sentence could be formed such as:
- By combining a noun phrase (NP) and a verb phrase (VP).
- By adding a conjunction between two sentences.

We show the concrete syntax for the above two functions along with some explanation:

```
lin UsePastS np vp =
  { s = np.s ! Nom ++ vp.s ! Past ! np.p ! np.n ! np.g } ;
```

This linearization rule states that the nominative form of noun phrase could be combined by the verb phrase (which is a past tense auxiliary in this case) and they both must agree for their Person, Number and Gender parameters. e.g. (ye mera qələm tʰa, یہ میرا قلم تھا, It was my pen)

```
lin UsePresS np vp = { s =
  np.s! Obl ++ "کو" ++ vp.s! Present! np.p! np.n! np.g};
```

Similarly this is the linearization rule for one of the functions responsible for building sentences having a noun phrase (which is a Pronoun + Postposition) and a verb phrase (which is a Verb + Auxiliary). e.g. (is ko kɪtɑbeṇ leni heṇ, اِس کو کتابیں لینی ہیں, He/she suppose to take the books).

In a similar fashion, we have implemented the noun phrases and verb phrases. We show some of the implemented rules for them below:

**DemPron → Num → CN → NP**[1] e.g. (ye do kɪtɑbeṇ, یہ دو کتابیں, these two books), (wo aik kɪtɑb, وہ ایک کتاب, that one book) etc
**DemPron→ PN → NP** e.g. (wo Ali, وہ علی, that Ali) etc
**NP → PostP → CN → NP** e.g. (is ko kɪtɑbeṇ, اِس کو کتابیں, to him the books) etc

**Verb_Aux → VP** e.g. (heṇ, ہیں, are) etc
**Verb → Verb_Aux → VP** e.g. (leni tʰiṇ, لینی تھیں, was suppose to take) etc

GF follows Interlingua-based approach. Hence for an Abstract syntax, we may provide Concrete syntax for different languages and GF can not only parse them but also translate them from one syntax (Concrete) to another; hence providing translation.

## 8 A Complete Example

We demonstrate the analysis of the following sentence as a complete example:
(is ko kɪtɑbeṇ leni heṇ) اِس کو کتابیں لینی ہیں
Transliteration: a(i)s kw ktabyN lyny hyN
Morphological analysis:

<اِس, a(i)s>

yih_66. یہ +DemPron - Sg Obl - Pers3_Near
mayN_68. مَیں +PersPron - Sg Pers3_Near Obl-
<کو, kw>
kw_18. کو +PostP -
<کتابیں, ktabyN>
ktab_824. کتاب +N - Pl Nom - Fem
<لینی, lyny>
lyna_2. لینا +Verb - Inf_Fem -
<ہیں, hyN>
hwna_0. ہونا +Verb_Aux - Present Pers1 Pl Masc -
hwna_0. ہونا +Verb_Aux - Present Pers1 Pl Fem -
....

Syntactic parsing:

UsePresS (UseNP (UsePron mayN_68) kw_18 (UseN ktab_824)) (UseVP lyna_2 hwna_0)

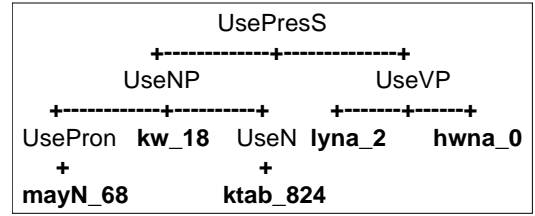

**Figure 2:** Syntax tree

## 9 Related Work

A large-scale on-going implementation of the Urdu grammar is the Parallel Grammar project (Butt & King, 2002). In this project, the Urdu/Hindi morphology is based on Xerox finite state technology and it relies on ASCII transliteration.

The Urdu Localization Project is also on-going project (Hussain, 2004). Its translation component is based on LFG formalism.

A number of publications are available for the above mentioned projects but their implementation is not publicly available.

EMILLE was a three year project in which 97 million word corpus was generated for the South Asian languages. For Urdu, an automated part-of-speech tagger was further developed (Hardie, 2005) that was then subsequently used to tag the Urdu corpus.

The CRL Language Resource Project[2] provides an Urdu Resource Package that contains an online Urdu-English dictionary and a morphological analyzer. However the design decisions regarding

---

[1] Type safe linguistic categories, (Num: Numerals, CN: Common Noun, DemPron: Demonstative Pronoun, VAux: Verb Auxiliary, PostP: Postposition)

[2] http://crl.nmsu.edu/Resources/lang_res

morphological implementation are not well documented.

A notable transliteration system for Urdu and Hindi is Abbas Malik's Hindi-Urdu Machine Transliteration System (Malik, 2006), in which SAMPA transcription System is used.

## 10 Results

FM has many merits and strengths for the development and implementation of a linguistic model and is proved to be a good choice for implementing Urdu morphology. Haskell provides us complete freedom for defining Urdu morphology with great ease. Dealing word classes and their parameters as algebraic data types, and the inflection tables (paradigms) for all word classes as finite functions satisfying the completeness, makes this implementation elegant, modular, extensible and reusable. We demonstrated the usability of this work by implementing a fragment of Urdu syntax in GF.

However, we do not provide a fully vocalized lexicon which is a fundamental limitation. Further, for the moment, for analysis of words, the runtime system of FM requires an exact match of a word or its word forms. Therefore one cannot check if there exist any orthographically different versions of a word in the lexicon.

## 11 Conclusions and Future work

This work presents an understanding of the Urdu language (morphology + orthography + lexicon) as well as a simple and straight-forward solution. Urdu is a challenging language and FM adequately fulfills it with a good margin.

This project could be further enhanced with the following possible extensions:
- A component that matches the partly vocalized input words with the canonical words in the lexicon, possibly returning multiple results.
- Algorithms that add the missing diacritics on partly vocalized words automatically.
- The remaining less frequent, very irregular group of words (especially loan Arabic and Persian words) in the inflection engine and a bigger coverage of lexicon.
- A comprehensive implementation for Urdu syntax.
- This system can equally be used for Hindi (morphology + syntax) by providing a lexicon and a transliteration scheme for Davanagari script.